# Delivery Line Tracking Robot

Md Rakibul Karim Akanda, Jason Lazo, Quintwon Carter, and Haineef Roberts

Department of Engineering Technology, Savannah State University—Savannah, GA 31404, United States of America

**Abstract**

The project we embarked on is making an electronic robot that can deliver a package along a set route through infrared sensors. It uses the infrared sensors to determine if the path it is following is correct or if it is off course. This is determined by sending off a photon to reflect off the path and determines if it is on a light surface by the amount of light emitted back or if it is a dark surface by the amount of light that is not present. In addition to following a line, the user can stop and start the robot at any interval through the infrared remote control. The project is a combination of the practical parts of machinery with the software part of coding in Arduino which is a coding subsect of C++. This can lead to endless possibilities that could help a wide variety of people from all ranges of life, especially with those that live with disabilities.

1. Introduction

Automation in daily life can be something that allows such difficult tasks to become a wanton action that anybody can do. There are a wide range of purposes that the robot delivery system can fill in practical daily life. This is a combination of compassion and forward thinking towards the future. As people age their mobility tends to decrease with time because of injuries due to falling [1]. Falling is an issue that can be problematic to even the point of deadly as a person progresses in age. It is shown that throughout the years, coming up to the age of a typical senior citizen, means that a person is more likely to suffered death instead of just an injury.

The automation of robots for delivering packages could greatly aid senior citizens and those with disabilities as it relates to physical mobility. These individuals may face difficulties leaving their homes to run errands or receive deliveries due to limitations in their physical ability, which can impact their independence and quality of life. With the use of robots for package delivery, these individuals can receive their packages without having to leave their homes, making their lives easier and more comfortable.

One of the benefits of using robots for package delivery is that they can navigate obstacles and uneven surfaces that might pose a challenge for individuals with mobility issues. This means that the robot can deliver packages directly to the doorstep of the recipient, regardless of the challenges presented by stairs or steep inclines. About 36 million falls are reported among older adults each year—resulting in more than 32,000 deaths [2]. Additionally, the use of robots for package delivery eliminates the need for individuals with mobility issues to carry heavy packages, which can be especially challenging for those with physical limitations.

Moreover, the automation of robots for package delivery could provide an extra layer of safety and security for senior citizens and those with disabilities. These individuals may be at a higher risk of theft or harm when out in public, particularly if they are carrying valuable items or are visibly struggling with mobility issues. With the use of robots for package delivery, the individual can receive their package without leaving their home, reducing the risk of exposure to potential harm or theft.

Overall, the automation of robots for package delivery could greatly improve the quality of life for senior citizens and those with disabilities as it relates to physical mobility. By providing a safe, secure, and convenient means of package delivery, individuals with mobility issues can receive the items they need without having to leave their homes or face physical challenges associated with package delivery.

This is not only an issue facing the older generation but for a large swath of people across the country. As stated by the Center of Disease Control 11% of all adults in the United States face an issue with mobility, such as walking or climbing stairs. This problem is only getting more exasperated as the general age of the population increases over time. The issue of how society maintains normalcy as aging becomes the norm, but people need to be as they once were. To put the how much of a tidal wave of that this issue is something that needs to be concerned about it is estimated that more than one out of every six, including children and younger, is 65 or older and only continues to get bigger as time goes forward [3].

2. **Equipment**

The project revolves around the programmable circuit board, also referred to as a microcontroller, the Arduino UNO. The flexibility of a microcontroller enables the user to assign inputs and outputs and control with a few lines of code. The equipment list is below.

| Component | Image |
|---|---|
| Arduino UNO (x1) | 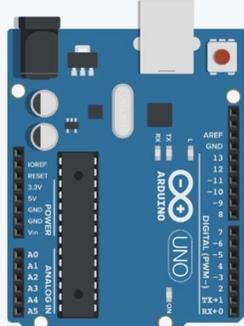 |
| Battery 9 Volt (x1) | 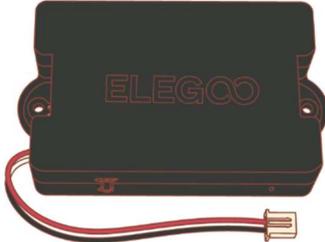 |
| Black Tape | 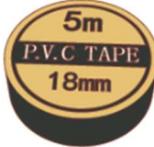 |
| Chassis | 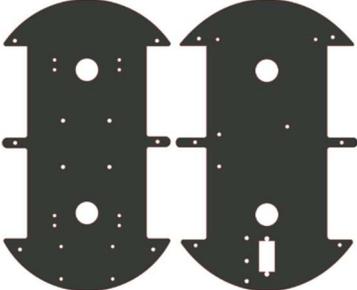 |
| DC Motor (x4) | 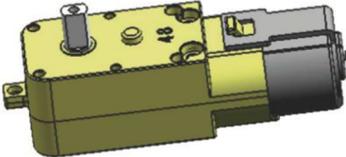 |
| Hardware (Assorted) |  |
| H-Bridge (TB6612 x1) |  |
| LED (Delivery x1) |  |
| Line Tracking Sensor (ITR20001 x3) | 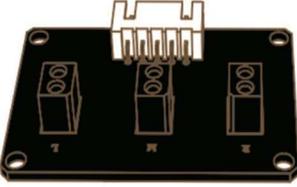 |
| Infrared Receiver (x1) | 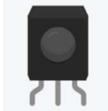 |
| Infrared Remote (x1) | 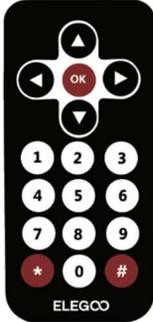 |
| IO Expansion Board (x1) | 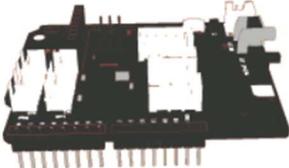 |
| Package Mechanism | 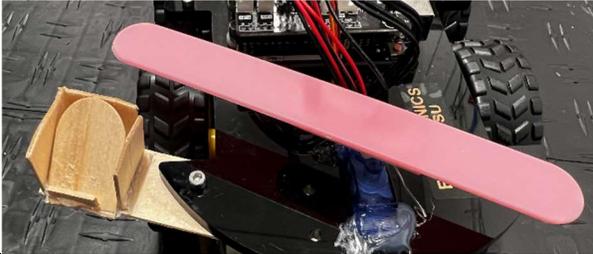 |
| Resistor | 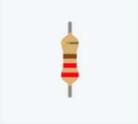 |
| Servo Motor (x1) | 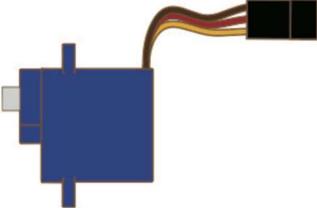 |
| Wheels (x4) | 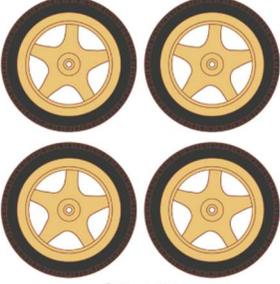 |

*[Table 1]*

*Arduino UNO*

A low-cost microcontroller board that can be programmed based on the C++ computer language. Arduino is an open-sourced accessible solution for a wide range of electronics, robotics, and programming projects.

*DC Motor*

Direct current motors have a large use base, anything from large industrial equipment to small toys. A typical DC motor has a pair of permanent magnets in the stator creating an electromagnetic field when current passes through the coil an electromagnetic current is induced causing the coil to rotate. The direction of the current will determine the direction of the armature to rotate. With many use cases, the application for a servo motor can be used at a vast scale.

*H-Bridge (TB6612)*

The TB6612 H-Bridge is used to drive higher amperage than capable with the Arduino microcontroller board. In so requiring a greater external power source. The H-Bridge consists of four switches that control voltage which allows the ability to apply voltage across the DC motors in different directions to rotate in either clockwise (CW) or counterclockwise (CWW) configurations.

*Line Tracking Sensor (ITR20001 x3)*

The line tracking module is composed of three ITR20001 photoelectric sensors, each perspective component is aligned for a Right, Center and Left sensor. A photoelectric sensor is a device useful to detect the presence or absence of an object using light. Each sensor consists of a phototransistor and an infrared light-emitting diode (LED). Detecting the light reflected from the LED sensor can detect the difference between light and dark lines.

*Infrared Receiver and Remote*

The IR receiver is sensitive to infrared light which detects the supplied infrared light from the IR remote at a certain frequency. The IR receiver is a photodiode that converts the IR signal light to an electrical signal. The IR remote is a transmitter with an IR LED that emits IR light. The encoded light signal from the remote is received to by the IR receiver and converted to binary for the Arduino microcontroller to intake. The binary signal is converted to hexadecimal and is viewable with the serial communication dialogue. Each button input from the IR remote has a unique code which can be assigned to a specific command. Buttons 1 and 2 were chosen to stop and resume all functions of the line tracking car.

*Servo Motor*

A simple servo motor consists of a DC motor combined with a potentiometer and a set of gears that can position the output shaft to a certain angle. Servo motors have angle limitations along with certain torque limits. Regardless of the limitations the reliability of a servo motor permanently established the popularity in robotics and other electronic applications.

3. **Theory**

The line following car displays a very simplified introduction to how real-world applications such as automated guided vehicles (AGV) and autonomous transportation work. Autonomous transportation technology navigates through various environments, including responding to obstacles, navigating on roads with different types of lines (or no lines at all), and encountering constantly evolving conditions. While automated guided vehicles follow along marked long lines or wires on the floor, or use radio waves, vision cameras, magnets, or lasers for navigation. The line following car is comprised of 4 key parts IR sensors, motor driver, an Arduino, and a power source. IR sensors are placed on both right and left side of the car, the car is situated over a black line such that both sensors are outside of the line. When the black line is detected by one of the sensors a signal is sent to the Arduino and the corresponding motor rotates backwards while the other motor rotates forward to turn the car to that side. This is theoretically comparable (but not identical) to how a real autonomous car might monitor lane lines to make sure the car does not drift out of its lane. There are many applications that display similar technology in this project such as automated guided vehicles in warehouse environments and autonomous transportation to name a few.

Automated guided vehicles are load carrying vehicles that navigate along the floor of a facility without guidance from personnel. Their movement is controlled by a mixture of software and sensor-based guidance systems. Since they move on a predetermined path along with accurately controlled acceleration and deceleration, and include automatic obstacle detection bumpers, AGVs provide safe movement of loads. AGV applications are used mainly in warehouses and manufacturing facilities with pallet trucks, forklifts, and transportation of light loads to name a few. There are various methods used to control AGV navigation, but for the sake of simplicity the 3 that will be covered are laser guided, guide tape, and wired (radio frequency). The guided tape navigation method is the method that closely relates to the method used in this project. The two styles of tape are colored and magnetic, with colored tape sensors sense the contrast between the tape and the background/floor to navigate and with magnetic tape there are inductive sensors that sense utilized to navigate.

AGV's provide many benefits such as safety, efficiency, and cost effectiveness. Safety is always the most important thing to any business AGVs can increase safety for personnel by performing tasks that could potentially dangers for humans along with the fact that they always follow their guide path and stop if they encounter an obstruction. Repetitive tasks can be completed with AGV's high volumes accuracy enabling employees to have more time to complete complicated tasks. High efficiency helps reduce unnecessary cost along with the reduced labor needs helps AGV's become a cost-effective solution.

**Circuit**

Additionally, a replica circuit of the project was created using 'Tinker Cad' online software simulator. The circuit was made as close as possible with the available components from the simulation program.

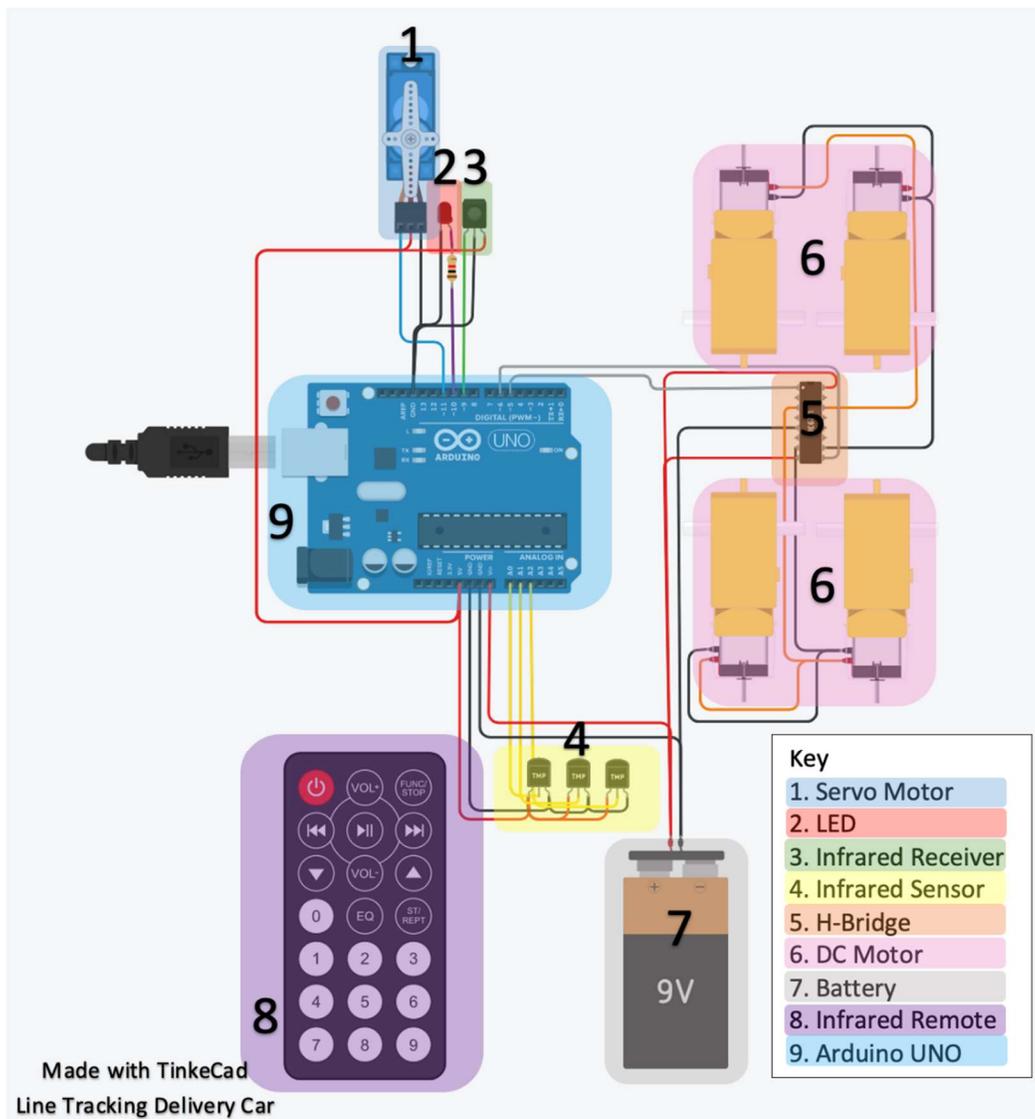

*Figure A*

(Please note the three ITR20001 infrared sensors used for the line tracking functions were substituted by the famous TMP36 sensors in Figure A, wiring is the same for both mentioned components).

The project relies heavily on the input from the line tracking system, as previously mentioned it is composed of three ITR20001 infrared sensors (Figure B). This sensor, similar to the Servo (Figure C) and Infared Receiver (Figure D) are utilizing a traditional three pin configuration made up of the 5 Volts supply, ground and their perspective signal wire.

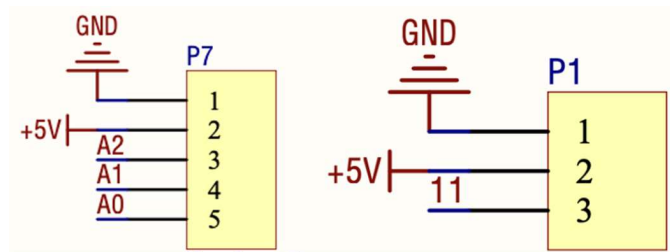

*Figure B*          *Figure C*

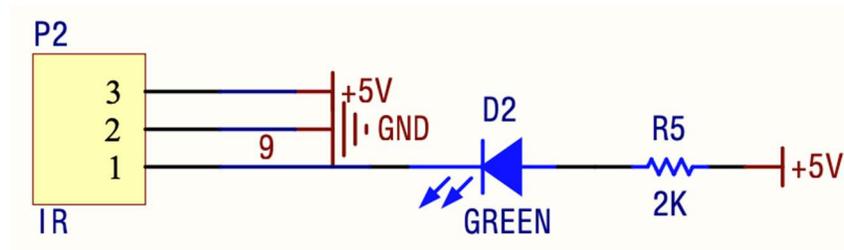

*Figure D*

The IR receiver Figure D is connected to PIN 9 on the Arduino UNO, an LED is connected in series with the input signal to illuminate a green LED when signal is received from the IR remote. A 2 kohm resistor is used as the small green LED does not require a vast amount of current. The expansion board supplied with the project kit does not specify the green LED current ratings but we can assume by using the supplied resistor with the following equation:

R = V / I  →  R = 5V – Vf / If  →  2k = 5 - 2 / If  →  If = 3 / 2k = 1.5 mA

In similar fashion, we can calculate the current limiting resistor on the red Delivery LED that was added to the project. The red Delivery LED is connected to PIN 10. The current limiting resistor was found using the following equation:

R = V / I   →   R = 5V – Vf / If   →   R = 5V – 2V / 20mA   →   R = 3V / 20 mA = 150 ohms

The decision was made to round up to a more common resistor value of 220 ohm. It is critical to round upwards to a greater robust resistor rating than to round down to prevent component failure from supply current.

Transitioning to the DC motors that propel the delivery car, the TB6612 H-Bridge is used to drive the load. The right-side motors are wired in series, and the left side are paired as well. The H-Bridge is critical to the operation of the delivery car as it grants us the ability to reverse the current in thus changing the direction of the wheels. The CW and CWW motion can be seen by reviewing the control function chart Figure F.

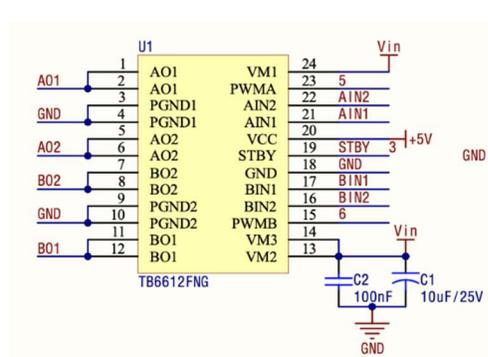

| Input | | | | Output | | |
|---|---|---|---|---|---|---|
| IN1 | IN2 | PWM | STBY | OUT1 | OUT2 | Mode |
| H | H | H/L | H | L | L | Short brake |
| L | H | H | H | L | H | CCW |
| | | L | H | L | L | Short brake |
| H | L | H | H | H | L | CW |
| | | L | H | L | L | Short brake |
| L | L | H | H | OFF (High impedance) | | Stop |
| H/L | H/L | H/L | L | OFF (High impedance) | | Standby |

*Figure E*                                    *Figure F*

The TB6612 is shown in Figure E, it is important to highlight the wiring function for this H-Bridge. For PIN 5, the right-side motors, are connected to PWMA which ultimately functions and an enable pin for the right side. The AIN1 signal is connected to PIN 7 which is our phase pin. AIN1 allows the motor to function forward/reverse (CW and CWW) depending on the 1/0 assignment in the code. The phase assignment was influential during turning movements as discussed in the Code section of this literature. The left-side motors also have the same structure with a PWMB in PIN 6 and a BIN1 phase in PIN 8.

**Code**

The code for this project is written in C++ language for the Arduino UNO microcontroller. See Figure G for the initial portion of the code. In this program we are utilizing two external libraries for the Servo Motor and the Infrared Remote and Receiver functions.

```
1   /* Arduino Line Follower Robot Code
2   by Jason Lazo, Quinton Carter, and Haineef Roberts */
3   #include <Servo.h>                  // Include the Servo Library
4   #include <IRremote.h>               // Include the IRremote library
5
6   #define IRReceiver 9                // Define the pin for the IR receiver module
7   IRrecv irrecv(IRReceiver);          // Create an IR receiver object
8   decode_results results;             // Create a variable to store the decoded results
9
10  #define Button1 0xFF6897            // Assinging IR HEX value to Button1
11  #define Button2 0xFF9867            // Assinging IR HEX value to Button2
12
13  //Line Tracker Module
14  #define L_S A2                      // Right Sensor
15  #define M_S A1                      // Middle Sensor
16  #define R_S A0                      // Left Sensor
17
18  #define DeliveryLED 10              // Delivery Alert LED
19
20  Servo DeliveryServo;                // Create servo object to control a servo
21
22  // H-Bridge Pins
23  #define ENABLE_RIGHT 5              // Right side motors  (PWMA)
24  #define ENABLE_LEFT 6               // Left side motors   (PWAB)
25  #define R_Direction 8               // Forward or Reverse (AIN1 : Phase Pin)
26  #define L_Direction 7               // Forward or Reverse (BIN2 : Phase Pin)
27  #define Master_Enable 3             // H-Bridge Enable    (STBY)
28
29  #define speed 100                   // Motor speed is 100 (0-255)
30  #define tspeed 120                  // Motor turning speed is 120 (0-255)
```

*Figure G*

Initial assignment of variables and pins.

Each button key pressed on the IR Remote generates a unique hexadecimal code. The hexadecimal code may vairy with different remotes. Utilizing the Serial.begin(9600) command line to initialize serial communication along with the irrecv.enableIRIn() line to start the IR receiver in Figure H, the ability to review the unique hexadecimals codes by our specific IR Remote. The baud rate of 9600 bits per second is more an enough to receive sufficient data to gather the button inputs. As shown in Figure G, hexadecimal 0xFF6897 and 0xFF9867 were gathered from Button 1 and 2 respectfully. A switch statement was used to specify which case was received for both Button 1 and Button 2 of the IR Remote.

Two different variables were created for the speed of DC Motors. Taking advantage of the PWM capabilities, a motor speed (0 to 255) was assigned for straight line motion and a different duty cycle for turning speed as the need for adjustment was found during testing. Ultimately the speed assignment is called upon the 'analogWrite' commands see Figure J and Figure K.

```
31
32  void setup(){
33
34    Serial.begin(9600);              // Initialize serial communication
35    irrecv.enableIRIn();             // Start the IR receiver
36
37    //Line Tracker Module (ITR20001)
38    pinMode(R_S, INPUT);             // Right IR diode
39    pinMode(M_S, INPUT);             // Middle IR diode
40    pinMode(L_S, INPUT);             // Left IR diode
41
42    //Delivery LED Alert
43    pinMode(DeliveryLED,OUTPUT);     // Assigning Delivery LED as output
44
45    //Motor Pin Setup
46    pinMode(ENABLE_RIGHT, OUTPUT);   // Right Motors (PWMA)
47    pinMode(ENABLE_LEFT, OUTPUT);    // Left Motors. (PWMB)
48    pinMode(R_Direction, OUTPUT);    // Right Motors Direction, high moves forwards
49    pinMode(L_Direction, OUTPUT);    // Left Motors Direction,  low moves backwards
50    pinMode(Master_Enable, OUTPUT);  // H-Bridge Enable
51
52    //Presets with Master Enable
53    digitalWrite(Master_Enable, HIGH);  // Enabling H-Bridge
54    digitalWrite(R_Direction, HIGH);    // Right Motors Direction, high moves forwards
55    digitalWrite(L_Direction, HIGH);    // Left Motors Direction,  low moves backwards
56
57    //Servo setup
58    DeliveryServo.attach(11);           // Attaches the servo on pin 11
59  }
```

*FIGURE H*

The main void loop of the code is mainly comprised of situations encountered by the line tracking module. Since this project is primarily dependent on the black line for output movements, the decision was made to create and call upon a function for each scenario the design may encounter.

```
61  void loop(){
62
63    DeliveryServo.write(0);              // Set the Delivery Servo to 0 degrees
64
65    if (irrecv.decode(&results))         // If IR signal is received
66    {
67        Serial.println(results.value, HEX);  // Print the received signal value
68        switch (results.value)               // Map the signal to the LED
69        {
70          case Button1:                      // Button1
71            IRGo();                          // IR Go function
72            break;
73          case Button2:                      // Button2
74            IRStop();                        // IR Stop function
75            break;
76        }
77        irrecv.resume();                     // Receive the next signal
78    }
79  // Line Tracking :  1 = black line :  0 = white background
80  // Forward Movment
81  // If Right Sensor and Left Sensor are at White color then it will call forword function
82    if((digitalRead(R_S) == 0)&&(digitalRead(L_S) == 0)&&(digitalRead(M_S) == 1)){forward();} \
83
84    // Right Turn
85    // If Right Sensor is Black and Left Sensor is White then it will call turn Right function
86    if ((digitalRead(L_S) == 0)&&(digitalRead(M_S) == 0)&&(digitalRead(R_S) == 1)){turnRight();delay(10);forward();}
87    if ((digitalRead(L_S) == 0)&&(digitalRead(M_S) == 1)&&(digitalRead(R_S) == 1)){turnRight();delay(10);forward();}
88
89    // Left turn
90    // If Right Sensor is White and Left Sensor is Black then it will call turn Left function
91    if ((digitalRead(L_S) == 1)&&(digitalRead(M_S) == 0)&&(digitalRead(R_S) == 0)){turnLeft();delay(10);forward();}
92    if ((digitalRead(L_S) == 1)&&(digitalRead(M_S) == 1)&&(digitalRead(R_S) == 0)){turnLeft();delay(10);forward();}
93
94    // Delivery Stop
95    if((digitalRead(R_S) == 1)&&(digitalRead(M_S) == 1)&&digitalRead(L_S)==1){Deliver();}
96
97    // Stoping at all white background
98    if ((digitalRead(R_S) == 0)&&(digitalRead(M_S) == 0)&&digitalRead(L_S) == 0){Stop();}
99  }
```

*FIGURE I*

```
100
101  void IRStop()                          //IR funtion to Stop
102  {
103      digitalWrite(DeliveryLED,HIGH);    // Turn Delivery LED on
104      digitalWrite(Master_Enable , LOW); // Disable H-Bridge
105  }
106
107  void IRGo()                            // IR function to Go
108  {
109      digitalWrite(DeliveryLED,LOW);     // Turn Delivery LED off
110      digitalWrite(Master_Enable , HIGH);// Enable H-Bridge
111  }
112
113  void forward(){ //forword
114      analogWrite(ENABLE_RIGHT, speed);  // Motor speed is 100 (0-255)
115      analogWrite(ENABLE_LEFT, speed);   // Motor speed is 100 (0-255)
116      digitalWrite(R_Direction, HIGH);   // Right Motor, high moves forwards
117      digitalWrite(L_Direction, HIGH);   // Left Motor,  low is moves backwards
118  }
119
120  void Stop(){ //stop
121      digitalWrite(ENABLE_RIGHT, LOW);   // Disabling Right Motors
122      digitalWrite(ENABLE_LEFT, LOW);    // Disabling Left Motors
123  }
124
125  void turnRight(){                      //turnRight
126      analogWrite(ENABLE_RIGHT, tspeed); // Motor turning speed is 120 (0-255)
127      analogWrite(ENABLE_LEFT, tspeed);  // Motor turning speed is 120 (0-255)
128      digitalWrite(R_Direction, HIGH);   // Right Motor, high moves forwards
129      digitalWrite(L_Direction, LOW);    // Left Motor,  low is moves backwards
130      //delay(10);
131  }
```

```
132
133  void turnLeft(){                       //turnLeft
134      analogWrite(ENABLE_RIGHT, tspeed); // Motor turning speed is 120 (0-255)
135      analogWrite(ENABLE_LEFT, tspeed);  // Motor turning speed is 120 (0-255)
136      digitalWrite(R_Direction, LOW);    // Right Motor, low is moves backwards
137      digitalWrite(L_Direction, HIGH);   // Left Motor,  high moves forwards
138      //delay(10);
139  }
140
141  void Deliver(){                        // Deliver
142      digitalWrite(Master_Enable, LOW);  // H-Bridge in standby
143      DeliveryServo.write(0);            // Setting servo to 0 degrees
144
145          for (int t = 0; t < 5;t++)    // Delivery LED blinking for delivery
146          {
147          digitalWrite(DeliveryLED, HIGH);
148          delay(200);
149          digitalWrite(DeliveryLED, LOW);
150          delay(200);
151          }
152
153      DeliveryServo.write(160);          // Setting servo to 160 degrees
154      delay (1000);
155
156      digitalWrite(Master_Enable, HIGH); // Enabling H-Bridge
157      digitalWrite(R_Direction, HIGH);   // Right Motor, high moves forwards
158      digitalWrite(L_Direction, HIGH);   // Left Motor,  low moves backwards
159  }
160
```

*Figure J*                                                     *Figure K*

As shown in Figure J and K, seven different functions were created to orchestrate the movements of the delivery car. The functions are as follows: Infrared Go and Infrared Stop to be executed via the inputs of the IR Remote. The remaining five functions are situational based from the black line path in front of the car. It needs to be mentioned that since the functions command movement the car will constantly be breaking in and out of the function statements as it travels the delivery path. This allows for the functions to be kept simple without the need for a complicated code structure to constantly break out of the current movement function. Although the system is not perfect, we are aware that the occasional requirement of human intervention is to physically reset the car back on delivery path.

### 4. Conclusion

This design is the meshing of two elements, hardware, and software, that are critical to the completion of real-world assignments where software and hardware developers work hand in hand together. Research on a variety of materials and tools that aid in the creation of tiny chips utilized in a variety of applications has been conducted during the past few decades [4-17]. In this project we replicated autonomous robotics used in industrial applications to perform repetitive delivery tasks with the line following car. During prototyping and the implementation of this project valuable troubleshooting skills and programming techniques were further developed. One of the major challenges during the prototyping was designing a mechanism that was sturdy enough for holding an item while being light enough to be within the torque range for the servo motor to actuate it, but success was achieved through dedicated group work. When designing any system there is a fair amount of troubleshooting involved to verify the correct operation of the system. Coding presented more troubleshooting opportunities followed by the electronic devices; this is nothing short of the activities that ensue the installation of the corresponding practical applications for this project. Throughout the course of this project the complete understanding of the line following car helps us gain insight into how closely the conceptualization of this project aligns with AGV's and AMR's (automated mobile robots). A complete understanding of these technologies helps us visualize the benefits that these solutions offer.